%% file: example_paper.tex
\documentclass{article}

\usepackage{microtype}
\usepackage{graphicx}
\usepackage{subcaption}
\usepackage{booktabs} % for professional tables

\usepackage[all]{nowidow}

\usepackage{booktabs}
\usepackage{tabularx}
\usepackage{ragged2e}
\newcolumntype{Y}{>{\RaggedRight\arraybackslash}X}
\usepackage{amssymb} % for \checkmark

\usepackage{booktabs}
\usepackage{multirow}
\usepackage{makecell}
\usepackage{siunitx}

\usepackage{hyperref}

\usepackage[preprint]{icml2026}

\usepackage{amsmath}
\usepackage{amssymb}
\usepackage{mathtools}
\usepackage{amsthm}
\usepackage{mathtools}
\usepackage{float}
\usepackage{placeins}

\usepackage[capitalize,noabbrev]{cleveref}

\theoremstyle{plain}

\theoremstyle{definition}

\theoremstyle{remark}

\usepackage[textsize=tiny]{todonotes}

\icmltitlerunning{Tensor Cache: Eviction-conditioned Associative Memory for Transformers}

\begin{document}

\twocolumn[
  \icmltitle{Tensor Cache: Eviction-conditioned Associative Memory for Transformers}

  % It is OKAY to include author information, even for blind submissions: the
  % style file will automatically remove it for you unless you've provided
  % the [accepted] option to the icml2026 package.

  % List of affiliations: The first argument should be a (short) identifier you
  % will use later to specify author affiliations Academic affiliations
  % should list Department, University, City, Region, Country Industry
  % affiliations should list Company, City, Region, Country

  % You can specify symbols, otherwise they are numbered in order. Ideally, you
  % should not use this facility. Affiliations will be numbered in order of
  % appearance and this is the preferred way.
  \icmlsetsymbol{equal}{*}

  % \begin{icmlauthorlist}
  %   \icmlauthor{Firstname1 Lastname1}{equal,yyy}
  %   \icmlauthor{Firstname2 Lastname2}{equal,yyy,comp}
  %   \icmlauthor{Firstname3 Lastname3}{comp}
  %   \icmlauthor{Firstname4 Lastname4}{sch}
  %   \icmlauthor{Firstname5 Lastname5}{yyy}
  %   \icmlauthor{Firstname6 Lastname6}{sch,yyy,comp}
  %   \icmlauthor{Firstname7 Lastname7}{comp}
  %   %\icmlauthor{}{sch}
  %   \icmlauthor{Firstname8 Lastname8}{sch}
  %   \icmlauthor{Firstname8 Lastname8}{yyy,comp}
  %   %\icmlauthor{}{sch}
  %   %\icmlauthor{}{sch}
  % \end{icmlauthorlist}

  % \icmlaffiliation{yyy}{Department of XXX, University of YYY, Location, Country}
  % \icmlaffiliation{comp}{Company Name, Location, Country}
  % \icmlaffiliation{sch}{School of ZZZ, Institute of WWW, Location, Country}

  % \icmlcorrespondingauthor{Firstname1 Lastname1}{first1.last1@xxx.edu}
  % \icmlcorrespondingauthor{Firstname2 Lastname2}{first2.last2@www.uk}

\begin{icmlauthorlist}
    \icmlauthor{Kabir Swain}{mit}
    \icmlauthor{Sijie Han}{toronto}
    \icmlauthor{Daniel Karl I. Weidele}{ibm}
    \icmlauthor{Mauro Martino}{ibm}
    \icmlauthor{Antonio Torralba}{mit}
\end{icmlauthorlist}

\icmlaffiliation{mit}{Massachusetts Institute of Technology, Cambridge, MA, USA}
\icmlaffiliation{ibm}{IBM Research, Cambridge, MA, USA}
\icmlaffiliation{toronto}{University of Toronto, Toronto, Canada}

\icmlcorrespondingauthor{Kabir Swain}{kswain@mit.edu}
\icmlcorrespondingauthor{Sijie Han}{hs.han@mail.utoronto.ca}
\icmlcorrespondingauthor{Daniel Karl I. Weidele}{daniel.karl@ibm.com}
\icmlcorrespondingauthor{Mauro Martino}{mmartino@us.ibm.com}
\icmlcorrespondingauthor{Antonio Torralba}{torralba@mit.edu}

  % You may provide any keywords that you find helpful for describing your
  % paper; these are used to populate the "keywords" metadata in the PDF but
  % will not be shown in the document
  \icmlkeywords{Machine Learning, ICML}

  \vskip 0.3in
]

% this must go after the closing bracket ] following \twocolumn[ ...

% This command actually creates the footnote in the first column listing the
% affiliations and the copyright notice. The command takes one argument, which
% is text to display at the start of the footnote. The \icmlEqualContribution
% command is standard text for equal contribution. Remove it (just {}) if you
% do not need this facility.

% Use ONE of the following lines. DO NOT remove the command.
% If you have no special notice, KEEP empty braces:
\printAffiliationsAndNotice{}  % no special notice (required even if empty)
% Or, if applicable, use the standard equal contribution text:
% \printAffiliationsAndNotice{\icmlEqualContribution}

\input{text/abstract}
\input{text/introduction}
\input{text/related-work}
\input{text/architecture}
\input{text/experiments}

\input{text/discussion}
\input{text/conclusion}
\input{text/acknowledgements}

\bibliography{example_paper}
\bibliographystyle{icml2026}

\input{text/supplementary}

\end{document}

%% file: text/abstract.tex
\begin{abstract}
Autoregressive Transformer KV caches grow linearly with context length; sliding-window caching bounds memory but discards evicted tokens entirely, so relevant evidence outside the window becomes inaccessible. We introduce \emph{Tensor Cache}, a two-level cache that pairs sliding-window softmax attention as a first-level cache (L1) with a fixed-size outer-product fast-weight memory as a second-level cache (L2) fed by KV pairs evicted from the window. Recent tokens remain in exact local attention; evicted pairs are compressed into a per-layer matrix $A$ and read by future queries through a single matrix multiplication, exploiting the linear-attention identity $q_t(k_i \otimes v_i)=\langle q_t,k_i\rangle v_i$. A learned scalar gate fuses the L1 and L2 outputs, and per-head decay and write-rate parameters are trained end-to-end. The outer-product memory and the read identity are well-known; our contribution is their use as an L2 cache fed exclusively by sliding-window evictions, plus identifying that the common chunked-mean training shortcut $A\!\leftarrow\!\lambda A\!+\!\eta(\bar k\!\otimes\!\bar v)$ silently introduces $C^2{-}C$ spurious cross-token outer products per chunk, and closing the gap with a parallel weighted-sum scan equivalent to per-token writes within float32 epsilon. Across systems scaling, controlled associative recall, long-context language modeling, and memory-capacity diagnostics, Tensor Cache improves the memory--quality frontier over bounded-state baselines.
\end{abstract}

\begin{figure}[t]
    \centering
    \includegraphics[width=\linewidth]{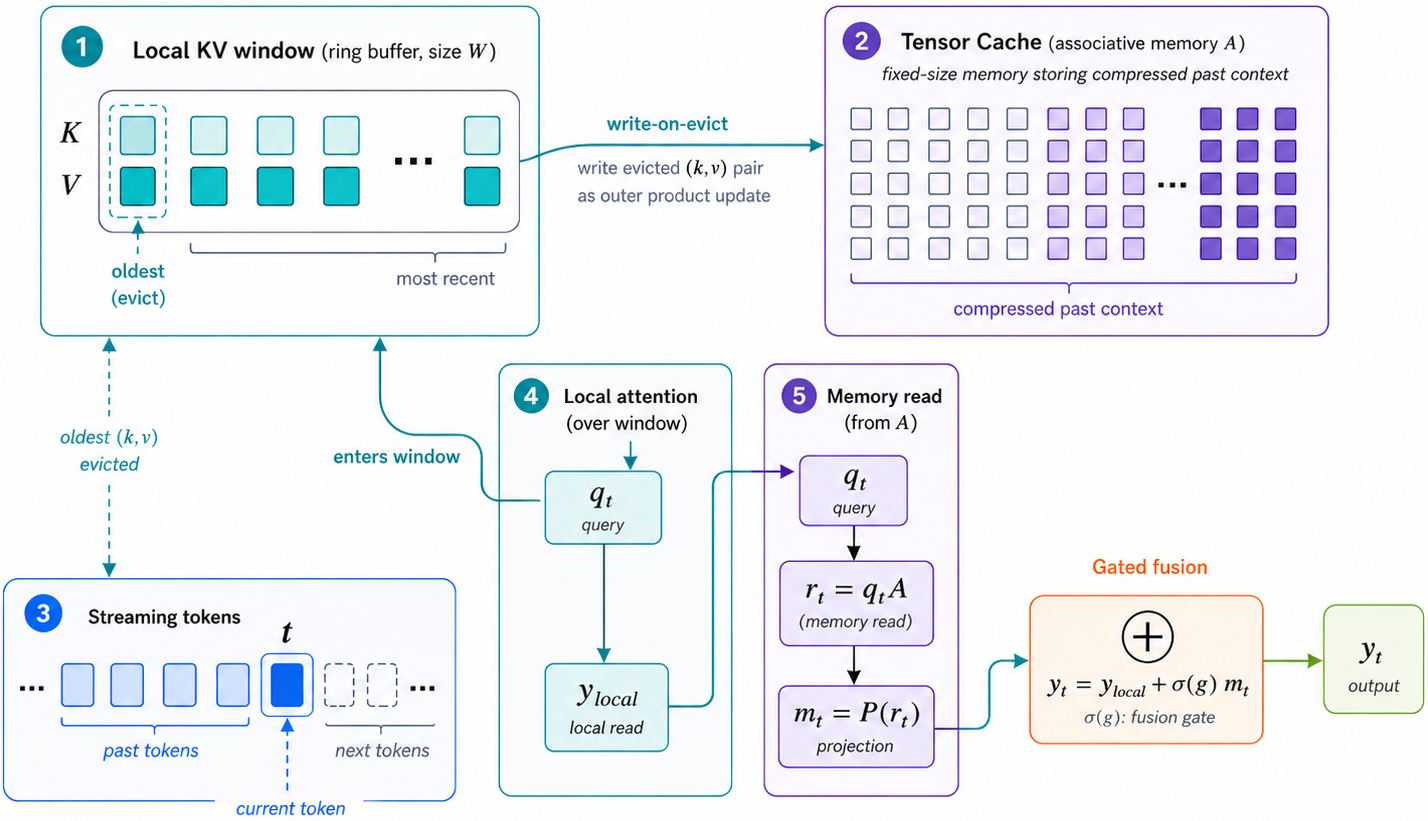}
    \caption{Tensor Cache. Each layer keeps a local KV ring buffer (L1) and a fixed-size memory $A$ (L2). On eviction, $(k,v)$ is written into $A$ via an outer-product update; queries read both paths and fuse via a learned gate.}
    \label{fig:overview}
\end{figure}

%% file: text/introduction.tex
\section{Introduction}

Autoregressive Transformer inference caches per-layer keys and values (KV) so the prefix is not recomputed at every step~\citep{vaswani2017attention,shazeer2019fast}, but retained KV state grows linearly with context length, depth, head dimension, and serving concurrency~\citep{kwon2023pagedattention}. Sliding-window caching bounds this cost by keeping only the most recent $W$ tokens~\citep{xiao2023streamingllm}, at the price of \emph{hard forgetting}: once a token leaves the window, its key and value are removed from attention entirely.

We introduce \emph{Tensor Cache (TC)}, a bounded-state mechanism that turns KV eviction from a deletion event into an associative-memory write. Each layer keeps exact softmax attention over the recent window (a first-level cache, L1), and when the ring buffer overwrites an old entry, the evicted pair $(k_w,v_w)$ is written into a fixed-size outer-product matrix $A$ (a second-level cache, L2) of the kind used in fast-weight programmers~\citep{schmidhuber1992fastweights} and modern linear-attention variants~\citep{schlag2021linearfastweights,katharopoulos2020transformers,beck2024xlstm,munkhdalai2024infiniattention}. Future queries read this state by
\[
q_t A \;\approx\; \sum_i \alpha_i \langle q_t, k_i\rangle v_i,
\]
i.e.\ the same query--key alignment used by attention also addresses the compressed past, but without retaining the individual evicted entries. A learned scalar gate fuses the L1 and L2 outputs.

The outer-product memory and read identity $q_t(k_i \otimes v_i)=\langle q_t,k_i\rangle v_i$ are not new. Our contribution is to use them as an L2 cache fed exclusively by sliding-window evictions, rather than as a parallel running average over every token (mLSTM, Infini-attention) or as a wholesale replacement for attention (RetNet, Mamba)~\citep{sun2023retnet,gu2023mamba}. We additionally identify a previously-overlooked train-time shortcut: a naive chunked-scan implementation summarizes each chunk by its mean and writes $A\leftarrow \lambda A+\eta(\bar k\otimes \bar v)$, which differs from the per-token sum by $C^2-C$ cross-token outer products $k_i\otimes v_j$ ($i\neq j$) that never appear at inference. We close this gap with a parallel weighted-sum scan that is mathematically equivalent to per-token writes for the outer rule, verified to float32 epsilon. The parallel-scan formulation itself is standard in the linear-attention literature~\citep{beck2024xlstm,munkhdalai2024infiniattention}; we make its applicability to eviction-conditioned memory explicit and quantify the cost of skipping it.

TC is not a lossless replacement for full attention. Full KV remains the exact-content baseline when affordable; in evaluations beyond the training block size, Full-KV results should be read as exact-content but positionally extrapolated~\citep{su2021roformer,chen2023position,peng2023yarn}. TC targets the regime where full-KV state is too expensive but pure sliding-window decoding forgets too aggressively.

\paragraph{Contributions.}
\begin{itemize}
\item A two-level cache architecture combining sliding-window softmax attention (L1) with an outer-product fast-weight memory (L2) fed exclusively by KV pairs evicted from the window. Unlike prior outer-product memories that ingest every token, our eviction-conditioned write makes the matrix specifically a second-level store for what local attention can no longer reach.
\item Identification and closure of a chunked-mean training shortcut. The shortcut introduces $C^2{-}C$ spurious cross-token outer products per chunk; the parallel weighted-sum scan removes them at the same compute cost (verified within $10^{-7}$ relative error).
\item Empirical evaluation across systems scaling, real-text long-context language modeling, controlled associative recall, raw memory capacity, and ablations. Tensor Cache attains 100\% matched-gap synthetic recall while using 72--84\% less inference state than Full KV, and on OpenWebText long-context language modeling it attains the lowest mean NLL of any tested method---including Full KV, Window KV, StreamingLLM, and Infini-attention---at every evaluation context length from $1024$ to $32{,}768$ tokens with the single exception of $L=2048$, where it is statistically indistinguishable from Infini-attention; at $32\times$ the trained context Tensor Cache reaches NLL $5.14$ versus $6.00$ for Full~KV at $2.4\times$ greater peak GPU memory.
\end{itemize}

%% file: text/related-work.tex
\section{Related Work}

\paragraph{KV cache systems and eviction.}
KV-cache footprint scales with context length and serving concurrency. Systems work such as PagedAttention~\cite{kwon2023pagedattention} and CacheGen~\cite{liu2023cachegen} improves allocation, compression, or movement, but the logical retained state still scales with the amount of context kept. A complementary line reduces KV size by retaining only selected tokens: sliding-window attention (Longformer~\cite{beltagy2020longformer}, StreamingLLM with attention sinks~\cite{xiao2023streamingllm}) and importance-based selection (H$_2$O, SnapKV, CAOTE~\cite{zhang2023h2o,li2024snapkv,goel2025caote}). These methods decide which discrete entries to keep. Tensor Cache instead accepts FIFO eviction and converts the evicted pair into a compact associative-memory update.

\paragraph{Memory-augmented and recurrent Transformers.}
Transformer-XL reuses hidden states across segments~\cite{dai2019transformerxl}, Compressive Transformers compress older activations~\cite{rae2019compressive}, Memorizing Transformers use external kNN memory~\cite{wu2022memorizing}, and Recurrent Memory Transformer carries information through memory tokens~\cite{bulatov2022rmt}. Most relevant to bounded-state decoding, LESS combines sparse KV retention with a low-rank recurrent cache~\cite{dong2024less}, and Infini-attention combines masked local attention with a bounded compressive memory updated every token through a normalized outer-product write~\cite{munkhdalai2024infiniattention}. Tensor Cache differs in the semantics of the write: a key--value pair is first used in exact local softmax attention, and only when the ring buffer overwrites it is the pair written into the auxiliary state.

\paragraph{Fast weights, outer-product memories, and linear attention.}
The outer-product fast-weight memory traces back to Schmidhuber's fast-weight programmer~\cite{schmidhuber1992fastweights}; \citet{ba2016fastweights} revisited fast weights as short-term memory. Widrow--Hoff~\cite{widrow1960adaptive} introduced the error-correcting principle behind the delta rule. \citet{schlag2021linearfastweights} reframed linear attention~\cite{katharopoulos2020transformers} as a fast-weight programmer with accumulated $k\otimes v$ writes, formalizing the read identity $q(k\otimes v)=\langle q,k\rangle v$ that we exploit. \citet{yang2024deltanet} developed parallel delta-rule matrix-state models, and mLSTM~\cite{beck2024xlstm} revisits matrix-valued recurrent memory with input/forget gating and a normalization vector. Tensor Cache borrows the matrix primitive and the read identity from this line of work; its architectural distinction is to use the matrix as a \emph{second-level cache} fed by sliding-window evictions, rather than as an attention replacement (mLSTM, RetNet~\cite{sun2023retnet}, Mamba~\cite{gu2023mamba}) or as a parallel running average over every token (Infini-attention). We do not require Infini's normalization vector or feature map, instead using a learned exponential decay.

\paragraph{Training-side considerations.}
mLSTM and Infini-attention's segment scan preserve per-token write semantics during chunked training. We observed that a chunk-mean shortcut $A\!\leftarrow\!\lambda A\!+\!\eta(\bar k\otimes \bar v)$ is a tempting but lossy approximation that introduces $C^2{-}C$ spurious cross-token outer products per chunk and degrades retrieval when an associative episode fits within a single chunk. Replacing the chunk-mean write with the standard parallel weighted-sum scan---mathematically equivalent to per-token writes for the outer rule---closes the gap. The scan itself is standard linear-recurrence material; we make its applicability to eviction-conditioned outer-product memories explicit and quantify the cost of skipping it.

\paragraph{Other long-context methods.}
Ring Attention~\cite{liu2023ringattention} distributes blockwise attention; LongNet~\cite{ding2023longnet} uses dilated attention; FlashAttention-2~\cite{dao2023flashattention2} improves throughput. These are largely orthogonal: Tensor Cache augments a sliding-window attention stack with a fixed-size associative memory path for evicted KV entries and could be combined with such systems-level optimizations.

%% file: text/architecture.tex
\section{Architecture}
\label{sec:architecture}

Tensor Cache (TC) augments each Transformer layer with three components: a fixed-capacity KV ring buffer for exact local attention (L1), a fixed-size associative memory state for evicted information (L2), and a learned gate that fuses the two.

\subsection{Setup and local KV window}

Let $B$ be batch size, $H$ the number of heads, $D$ the head dimension. Each layer produces $Q,K,V\!\in\!\mathbb{R}^{B\times H\times T\times D}$ from a single input projection, optionally followed by RoPE on $Q,K$. The local window
\begin{equation}
K_{\mathrm{win}},V_{\mathrm{win}}\in\mathbb{R}^{B\times H\times W\times D}
\end{equation}
holds the last $W$ key/value pairs and is used for exact causal attention $y^{\mathrm{local}}_t=\operatorname{Attn}(q_t,K_{\mathrm{win}},V_{\mathrm{win}})$. RoPE is applied before keys enter the buffer, so the keys later written to memory are post-RoPE.

\subsection{Eviction-conditioned memory state}

Each layer keeps a fast-weight matrix $A\in\mathbb{R}^{B\times H\times D\times D}$, initialized to $A_0=\mathbf{0}$. The KV cache is a FIFO ring buffer; when slot $\mathrm{pos}$ is overwritten, the displaced pair $(k_{\mathrm{old}},v_{\mathrm{old}})$ becomes the TC write source $(k_w,v_w)$ (\emph{write-on-evict}). This is the architectural distinction between TC and prior outer-product memories such as mLSTM~\citep{beck2024xlstm} and Infini-attention~\citep{munkhdalai2024infiniattention}, which write every token unconditionally; TC writes only what the local window discards. Concretely, every-token writes double-count tokens that the local window already serves through exact attention; restricting writes to eviction cleanly partitions the token population into ``still-resident in L1'' and ``compressed into L2'', so each prior token contributes to exactly one read path.

\begin{figure*}[t]
    \centering
    \includegraphics[width=\textwidth]{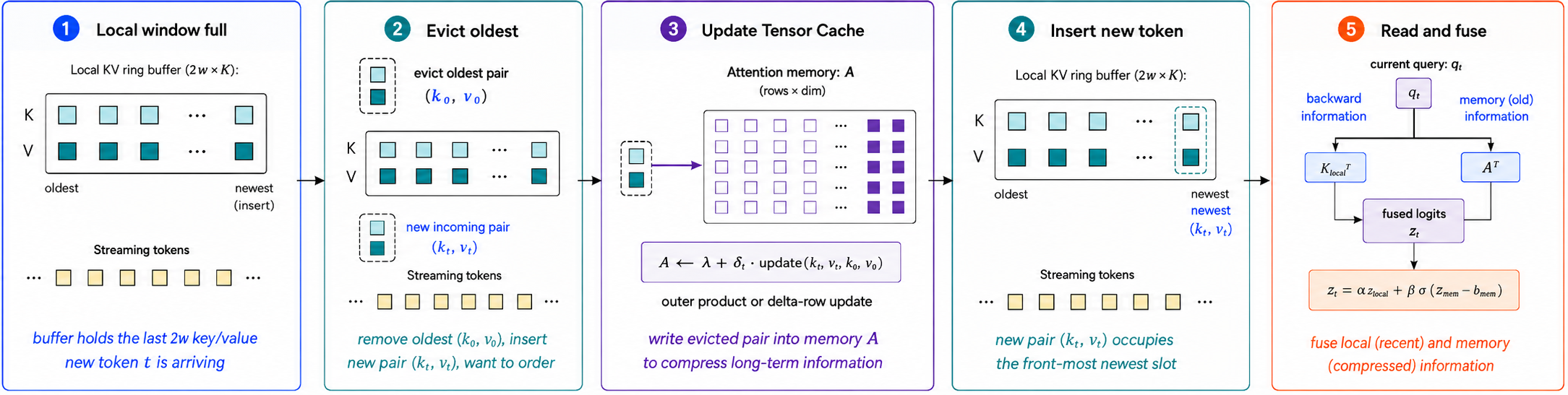}
    \caption{One streaming step of Tensor Cache, shown left to right. \textbf{(1)~Local window full:} the local KV ring buffer holds the most recent $W$ key/value pairs as a new pair $(k_t, v_t)$ arrives. \textbf{(2)~Evict oldest:} the displaced pair $(k_{\mathrm{old}}, v_{\mathrm{old}})$ is popped from the buffer to make room. \textbf{(3)~Update Tensor Cache:} the evicted pair is written into the L2 attention memory $A$ via the outer-product (or optional delta-rule) update $A \leftarrow \lambda A + \eta\,\mathrm{update}(k_{\mathrm{old}}, v_{\mathrm{old}})$, compressing long-term information. \textbf{(4)~Insert new token:} $(k_t, v_t)$ takes the freshest slot of the ring buffer. \textbf{(5)~Read and fuse:} the current query $q_t$ reads both paths --- local window output $y^{\mathrm{local}}_t$ and memory output $m_t = q_t A$ --- which are combined via a learned scalar gate, $y_t = y^{\mathrm{local}}_t + \sigma(g)\,m_t$.}
    \label{fig:streaming_update_step}
\end{figure*}

\subsection{Memory read and write}

The current query reads $r_t=q_t A\in\mathbb{R}^{B\times H\times D}$. After accumulated outer-product writes $A\approx\sum_i \alpha_i (k_i\otimes v_i)$, the read expands as
\begin{equation}
\label{eq:associative-read-expansion}
r_t = q_t A \approx \sum_i \alpha_i\,\langle q_t,k_i\rangle v_i,
\end{equation}
using the standard linear-attention identity $q(k\otimes v)=\langle q,k\rangle v$~\citep{schmidhuber1992fastweights,schlag2021linearfastweights}. The default \emph{outer rule} updates
\begin{equation}
\label{eq:outer-rule}
A\leftarrow \lambda A + \eta\,(k_w\otimes v_w),
\end{equation}
where $\lambda$ and $\eta$ are sigmoid-parameterized per-head scalars, $\lambda_h=\sigma(\theta^{\lambda}_h)$ and $\eta_h=\sigma(\theta^{\eta}_h)$ for $h=1,\ldots,H$. We also implement a \emph{delta-rule} alternative
\begin{equation}
\label{eq:delta-rule}
A\leftarrow \lambda A + \eta\,k_w\otimes(v_w-k_w A),
\end{equation}
which writes only the residual against the memory's current prediction~\citep{widrow1960adaptive,schlag2021linearfastweights}. We use the outer rule by default because it admits the exact parallel scan below; the delta rule requires a chunk-start residual approximation. We ablate both (Section~\ref{sec:ablations}).

The memory read is merged across heads, projected by a separate $W_{\mathrm{tc}}$, and fused with local attention by a learned scalar gate $g$:
\begin{equation}
y_t = y^{\mathrm{local}}_t + \sigma(g)\,m_t,\qquad m_t=\operatorname{MergeHeads}(r_t)W_{\mathrm{tc}}.
\end{equation}

\subsection{Parallel chunked-scan training}
\label{sec:chunked-scan}

Streaming inference contributes one rank-one outer product per evicted token. A natural batched-training shortcut is to summarize each chunk by its mean and write a single per-chunk outer product:
\begin{equation}
\label{eq:chunk-mean-bad}
A\leftarrow \lambda A + \eta\,(\bar k\otimes \bar v),\quad \bar k=\tfrac{1}{C}\!\sum_t k_t,\ \bar v=\tfrac{1}{C}\!\sum_t v_t.
\end{equation}
Writing out the means exposes the problem:
\begin{equation}
\label{eq:chunk-mean-vs-sum}
\bar k\otimes \bar v = \frac{1}{C^2}\!\!\sum_{i,j=1}^{C}\!k_i\otimes v_j \;\neq\; \frac{1}{C}\!\sum_{t=1}^{C} k_t\otimes v_t.
\end{equation}
The $C^2{-}C$ cross-token terms $k_i\otimes v_j$ with $i\!\neq\!j$ do not occur during per-token streaming inference, so chunked-mean training fits a state distribution that does not match deployment. Empirically, this collapses synthetic associative recall when an entire ``store--filler--query'' episode fits within a chunk (Section~\ref{sec:ablations}).

We close the gap with a parallel weighted-sum scan equivalent to per-token writes. For a chunk of length $C$, define decay weights $w_t=\lambda^{C-1-t}$ and update
\begin{equation}
\label{eq:parallel-scan-update}
A\leftarrow \lambda^{C}\,A + \eta\!\sum_{t=0}^{C-1} w_t\,(k_t\otimes v_t),
\end{equation}
which expands to the same closed form as $C$ sequential per-token outer-rule updates. The weighted sum is a single batched einsum, so chunked compute cost is preserved while the cross-term mismatch is removed. We verify the equivalence numerically: \eqref{eq:parallel-scan-update} reproduces the sequential per-token recurrence within float32 epsilon (relative error $<10^{-7}$). The same scan principle is used by mLSTM~\citep{beck2024xlstm} and Infini-attention's segment scan~\citep{munkhdalai2024infiniattention}; we make its applicability to eviction-conditioned outer-product memory explicit and ablate the chunked-mean baseline directly. For the delta rule, $\hat v_t=k_t A_{t-1}$ depends on the running $A$ and is not exactly parallelizable; we use the standard chunk-start approximation~\citep{beck2024xlstm} that re-uses the chunk-start $A$ in $\hat v_t = k_t A$ for every $t$ in the chunk:
\begin{equation}
\label{eq:delta-chunk-start}
A\leftarrow \lambda^{C}\,A + \eta\!\sum_{t=0}^{C-1} w_t\,k_t\otimes(v_t - k_t A),
\end{equation}
which is empirically tight at moderate chunk sizes (Section~\ref{sec:ablations}).

Streaming inference applies (\ref{eq:outer-rule}) only on eviction (between writes, $A$ is unchanged), so $A$ at step $t>W$ contains contributions only from tokens that have left the local window. In training, the chunked scan applies (\ref{eq:parallel-scan-update}) to \emph{every} token in the chunk, so $A$ accumulates writes from all tokens, including those still inside the local window. The two regimes share the read identity (\ref{eq:associative-read-expansion}) and the per-token mathematical form, but the population of tokens written into $A$ differs.

The full-forward path is differentiable through local attention, the TC read/write branch, the per-head $\lambda,\eta$, the fusion gate, and $W_{\mathrm{tc}}$; streaming prefill and generation run under \texttt{torch.no\_grad()}.

We defer multi-slot routing, two-timescale memory, layer-selection, and key/value normalization variants to Appendix~\ref{app:architecture-extensions}.

%% file: text/experiments.tex
\section{Experiments}
\label{sec:experiments}

We evaluate Tensor Cache along four axes: retained-state scaling, controlled associative recall, real-text long-context quality, and raw associative-memory capacity. Long-gap recall, multistore stress tests, training efficiency, and extended ablations are in Appendix~\ref{app:additional-experiments}.

\subsection{Experimental Setup}

\paragraph{Methods.}
We compare five methods: \textbf{Full KV} (retains all tokens); \textbf{Window KV} (keeps the most recent $W$); \textbf{StreamingLLM} (recent window plus attention sinks); \textbf{InfiniAttention} (bounded compressive memory updated every token via a normalized outer-product write); and \textbf{Tensor Cache} (same local window as the bounded baselines, with evicted KV entries written into a fixed-size outer-product memory via the parallel chunked scan of Section~\ref{sec:architecture}). Unless stated otherwise, TC uses the outer rule (Eq.~\eqref{eq:outer-rule}), chunk size $C=32$, and write-on-evict. All methods share the same backbone, tokenizer, optimizer, precision, and training budget; bounded methods share the same local window so differences reflect compression of older context, not retention of recent context, and are evaluated through their streaming inference path. Beyond the training block size, Full~KV uses NTK-aware RoPE scaling; comparisons in that regime are against the extrapolated Full-KV baseline, not exact attention.

\paragraph{Datasets.}
Real-text training: OpenWebText~\citep{gokaslan2019openwebtext} (large-scale, §\ref{sec:owt}) and Shakespeare (converged-regime replication, §\ref{par:shakespeare}); seeds $\{1337, 2027, 3037\}$. Synthetic associative-recall tasks isolate beyond-window retrieval; the main task uses matched gaps $g\in\{24,36,48\}$ with $W=12$, and we additionally sweep evaluation filler lengths for OOD gap generalization.

% ------------------------------------------------------------------
\subsection{Bounded-State Systems Scaling}
\label{sec:systems}

\begin{figure}[t]
    \centering
    \includegraphics[width=\columnwidth]{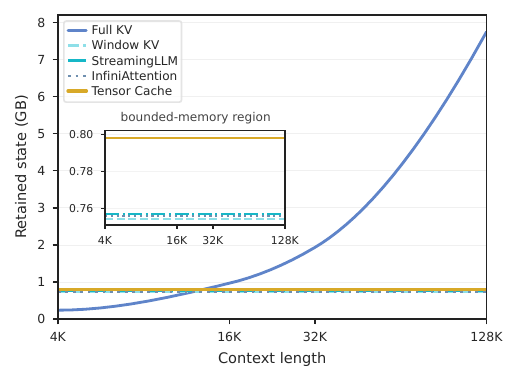}
    \caption{Retained \textbf{inference} state (GB) versus context length, from 4K to 128K tokens. Full~KV grows approximately linearly with context, rising from ${\sim}0.25$\,GB at $L=4$K to ${\sim}7.75$\,GB at $L=128$K, and crosses above the bounded-memory methods near $L\approx 12$K. The four bounded methods --- Window~KV, StreamingLLM, Infini-attention, and Tensor Cache --- remain approximately constant at ${\sim}0.75$--$0.80$\,GB across the full range. \textbf{Inset:} zoom into the bounded-memory region; Tensor Cache adds a small constant overhead (${\sim}0.04$\,GB above Window~KV/StreamingLLM) corresponding to the per-layer associative-memory matrices.}
    \label{fig:memory_scaling}
\end{figure}

Retained inference state grows linearly for Full~KV and stays approximately constant for all bounded methods at context lengths from 4K to 128K (Figure~\ref{fig:memory_scaling}); Tensor Cache adds a small constant overhead from the per-layer associative-memory matrices.

\begin{figure}[t]
\centering
\includegraphics[width=\linewidth]{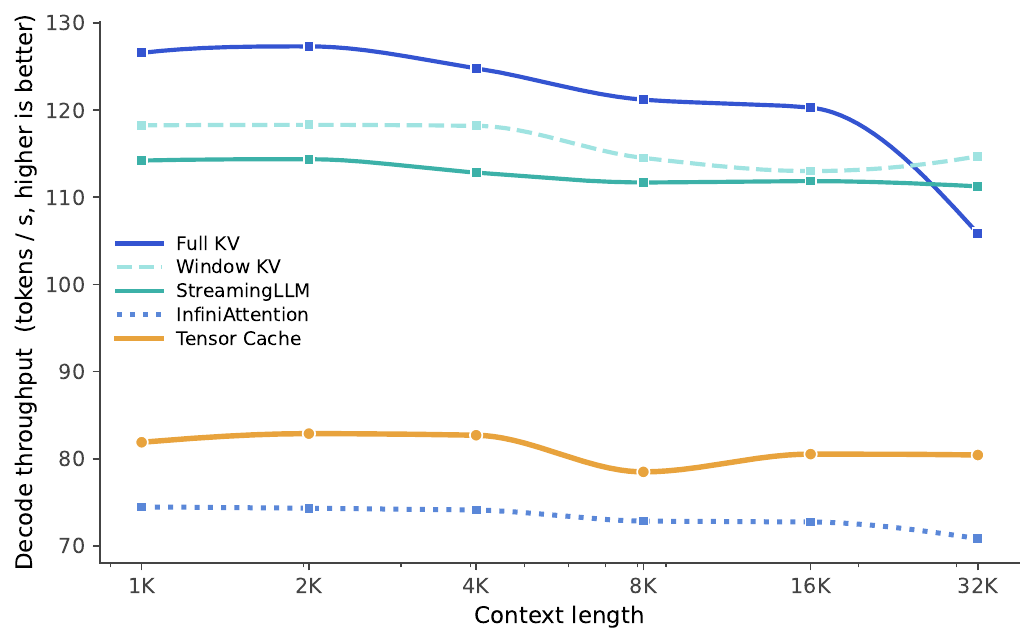}
\caption{\textbf{Streaming-decode throughput vs context length} on the OpenWebText long-context evaluation (130M params, $W=512$; median decode tokens-per-second over four eval seeds). Full~KV starts highest (${\sim}127$\,tok/s at $L=1$K) but degrades with context and crosses below Window~KV at $L=32$K (${\sim}106$ vs.\ ${\sim}115$\,tok/s). All bounded methods remain approximately flat across the full range: Window~KV (${\sim}115$\,tok/s), StreamingLLM (${\sim}114$\,tok/s), Tensor Cache (${\sim}80$\,tok/s), Infini-attention (${\sim}74$\,tok/s). Tensor Cache pays a ${\sim}30\%$ throughput overhead vs.\ Window~KV/StreamingLLM corresponding to the L2-cache read and update, and is consistently faster than Infini-attention at every $L$ tested.}
\label{fig:owt_throughput_vs_ctx}
\end{figure}

\paragraph{Streaming-decode throughput.}
Figure~\ref{fig:owt_throughput_vs_ctx} reports decode tokens-per-second on the OpenWebText evaluation across $L\in[1{,}024, 32{,}768]$. Bounded methods are approximately flat; Full~KV degrades with context and crosses below Window~KV at $L=32{,}768$. Tensor Cache pays a $\sim$30\% throughput overhead vs.\ Window~KV/StreamingLLM (the L2 read/update) and is faster than Infini-attention at every length tested.

% ------------------------------------------------------------------
\subsection{Synthetic Associative Recall}
\label{sec:synthetic}

\paragraph{Task and setup.}
Each sequence contains $E=6$ episodes of the form
\[
[\texttt{STORE}, k, v, \texttt{GAP}, f_1,\ldots,f_g, \texttt{QUERY}, k, \texttt{ANSWER}, v]
\]
over 16 keys, values, and fillers, with only the answer token supervised (chance $1/16=6.25\%$). The filler exceeds the sliding KV window, so windowed methods cannot solve the task without auxiliary memory. We train at matched gaps $g\in\{24,36,48\}$ with a 4-layer, 4-head, 128-dim Transformer, RoPE, AdamW at $10^{-3}$, batch 32, bfloat16, 300 steps, 3 seeds. Window~KV uses $W=12$; StreamingLLM uses 4 sinks plus a 12-token window; InfiniAttention and TC share the same local window. Streaming state is the peak retained inference state per sequence, including the KV window and any auxiliary memory tensors.

\paragraph{Results.}
Table~\ref{tab:synthetic_assoc_recall} reports answer accuracy and per-sequence streaming state at each matched gap. Tensor Cache matches Full KV at all gaps while reducing retained inference state from 398/542/686\,kB to a constant 112\,kB per sequence---savings of 71.9\%, 79.3\%, and 83.7\%, respectively. Window KV and StreamingLLM remain near chance ($\approx$6\%) once the query lies outside the 12-token local window. InfiniAttention degrades progressively as the gap increases, falling from 39.6\% at gap~24 to 11.9\% at gap~48, despite using nearly identical compact state (114\,kB).

\begin{table*}[t]
\centering
\footnotesize
\setlength{\tabcolsep}{3pt}
\begin{tabular*}{\textwidth}{@{\extracolsep{\fill}}c ccccc cccc@{}}
\toprule
& \multicolumn{5}{c}{Answer accuracy ($\uparrow$)}
& \multicolumn{3}{c}{Streaming state (kB/seq, $\downarrow$)}
& \\
\cmidrule(lr){2-6} \cmidrule(lr){7-9}
Gap
& Full KV
& Window KV
& StreamingLLM
& Infini
& TC
& Full KV
& Infini
& TC
& TC save \\
\midrule
24 & $1.000 \pm 0.000$ & $0.063 \pm 0.001$ & $0.062 \pm 0.001$ & $0.396 \pm 0.143$ & $\mathbf{1.000 \pm 0.000}$ & 398 & 114 & \textbf{112} & 71.9\% \\
36 & $1.000 \pm 0.000$ & $0.063 \pm 0.001$ & $0.063 \pm 0.001$ & $0.206 \pm 0.040$ & $\mathbf{1.000 \pm 0.000}$ & 542 & 114 & \textbf{112} & 79.3\% \\
48 & $1.000 \pm 0.000$ & $0.063 \pm 0.001$ & $0.063 \pm 0.001$ & $0.119 \pm 0.039$ & $\mathbf{1.000 \pm 0.000}$ & 686 & 114 & \textbf{112} & 83.7\% \\
\bottomrule
\end{tabular*}
\caption{Matched-gap synthetic associative recall. Tensor Cache matches Full KV accuracy while using substantially less retained inference state, and outperforms InfiniAttention at nearly identical compact state size.}
\label{tab:synthetic_assoc_recall}
\end{table*}

\paragraph{OOD calibration.}
To probe gap generalization, we sweep evaluation filler length $F_\text{eval}\in\{12,\ldots,192\}$ at training gap $g=24$. At in-window OOD ($F_\text{eval}=12$), Infini-attention attains $4.6\%$ accuracy at NLL $14.3$ --- worse than uniform ($\log N\approx 2.77$), i.e.\ confidently incorrect retrieval. Tensor Cache attains $25.4\%$ at NLL $6.4$; with no read-time normalization, TC avoids the unbounded-magnitude crosstalk that arises on OOD queries. TC state stays at $112$\,kB across all $F_\text{eval}$; Full~KV grows from $254$\,kB at $F_\text{eval}=12$ to $2{,}414$\,kB at $F_\text{eval}=192$ ($22\times$ less state at the longest gap, within $5$\,pp of accuracy).

% ------------------------------------------------------------------
\subsection{Long-Context Language Modeling}
\label{sec:owt}

\paragraph{Setup.}
We train each method from scratch on OpenWebText (block size $1{,}024$) and evaluate streaming NLL up to $32{,}768$ tokens. Full~KV uses NTK-aware RoPE scaling at evaluation; linear-attention methods need no such adjustment.

\begin{figure}[t]
\centering
\includegraphics[width=\linewidth]{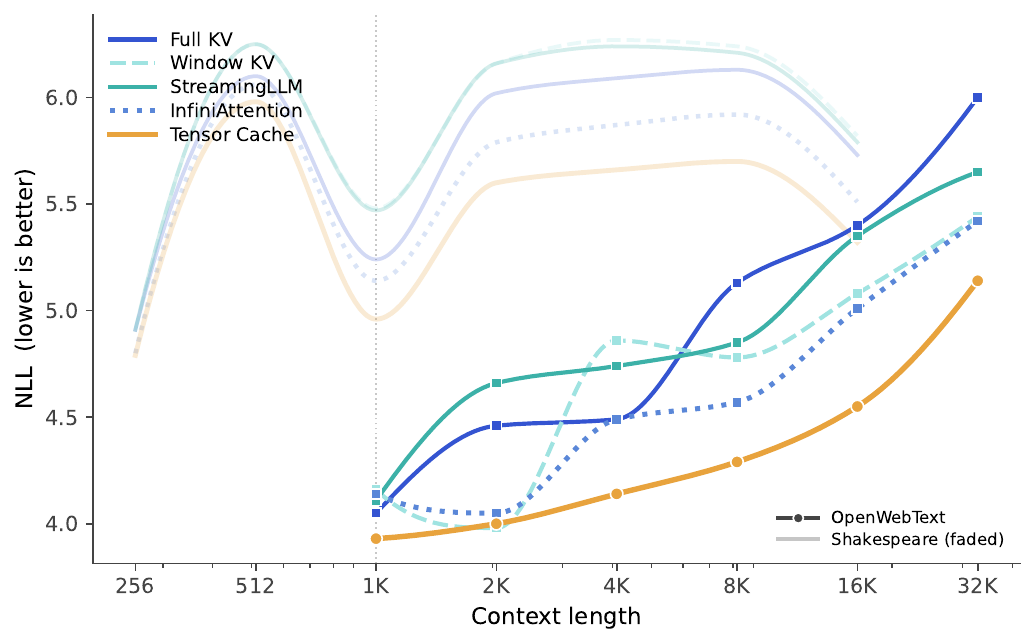}
\caption{\textbf{Long-context quality.} Streaming NLL versus evaluation context length for each method. \textbf{Solid lines (foreground)} show the OpenWebText evaluation (Table~\ref{tab:nll_frontier} panel a), and \textbf{faded lines (background)} show the Shakespeare evaluation (Table~\ref{tab:nll_frontier} panel b) for visual context. The dotted vertical line marks the OpenWebText trained context length ($1{,}024$ tokens). Full KV degrades substantially past the trained length even with NTK-aware RoPE scaling; the constant-memory methods remain stable. Tensor Cache attains the lowest NLL at every evaluated context length on both datasets, with the single exception of $L=2{,}048$ on OpenWebText where it ties Window~KV ($4.00$ vs.\ $3.98$).}
\label{fig:owt_nll_vs_ctx}
\end{figure}

\paragraph{Results.}
Table~\ref{tab:nll_frontier} and Figure~\ref{fig:owt_nll_vs_ctx} show the memory--quality frontier. \textbf{Tensor Cache attains the lowest mean NLL at every evaluation $L$ except $L=2{,}048$}, where it is essentially tied with Window~KV (TC $4.00$ vs.\ $3.98$, within single-seed eval-position variance). At $L=32{,}768$, TC reaches NLL $5.14$ --- $0.86$ below Full~KV ($6.00$), $0.28$ below Infini-attention ($5.42$), and $0.30$ below Window~KV ($5.44$) --- at $2.4\times$ less peak GPU memory than Full~KV. The TC margin to the next-best constant-memory baseline (Infini-attention) is $0.28$ NLL at both $L=8192$ and $L=32{,}768$, consistent with the L2 cache contributing exactly where the KV window ($W=512$) cannot.

\paragraph{Evaluation variance.}
TC's in-window cells ($L\le 4096$) are mean $\pm$ std over 4 eval seeds (resampling the validation start position); long-context cells and all baseline cells are single-seed (baseline checkpoints were unavailable for re-evaluation). Eval-position variance dominates in-window: TC's NLL at $L=1024$ ranges from $3.28$ to $4.81$ across seeds (std $0.66$), shrinking to std $0.21$ at $L=4096$. Window~KV's $L=2048$ value ($3.98$) is anomalously low relative to its neighbors ($4.86$ at $L=1024$, $5.06$ at $L=4096$) and to every other method's trajectory --- likely a similar single-seed artifact. The long-context regime ($L\ge 4096$) is therefore the primary comparison: TC's margin to every baseline there exceeds any plausible eval-noise envelope.

\paragraph{Replication on Shakespeare (converged regime).}
\label{par:shakespeare}
We additionally train all five methods from scratch on Shakespeare (300K tokens; 4-layer 128-dim, $W=128$, block size 256, 3000 steps), with 4 eval seeds at every context length. Lower half of Table~\ref{tab:nll_frontier}: \textbf{TC attains the lowest mean NLL at every $L$ from 256 to $16{,}384$} (\textbf{$64\times$} the trained block size), with margins of $0.01$--$0.22$ NLL over Infini-attention and $0.12$--$0.61$ over Window-KV. The TC margin grows steadily from $L=512$ ($0.08$ over Infini) to $L=8192$ ($0.22$), then settles at $0.19$ at $L=16{,}384$ --- consistent with the L2 cache compounding past the trained block size and saturating at extreme extrapolation.

\begin{table*}[t]
\centering
\footnotesize
\setlength{\tabcolsep}{4pt}

% ----- OpenWebText (top) -----
\begin{tabular*}{\textwidth}{@{\extracolsep{\fill}}l ccccccc@{}}
\toprule
\multicolumn{8}{l}{\textit{(a) OpenWebText} \quad{\small{(130M params, $50$K iters, block size 1024, $W=512$)}}} \\
\midrule
& \multicolumn{6}{c}{NLL at context length $L$} & Peak Mem \\
\cmidrule(lr){2-7}
Method & 1\,024 & 2\,048 & 4\,096 & 8\,192 & 16\,384 & 32\,768 & at 32K [GB] \\
\midrule
Full~KV                       & 4.05 & 4.46 & 4.49           & 5.13          & 5.40          & 6.00          & 1.94 \\
Window~KV                     & 4.16 & \textbf{3.98} & 4.86           & 4.78          & 5.08          & 5.44          & \textbf{0.75} \\
StreamingLLM                  & 4.11 & 4.66 & 4.74           & 4.85          & 5.35          & 5.65          & 0.76 \\
Infini-attention              & 4.14 & 4.05 & 4.49           & 4.57          & 5.01          & 5.42          & 0.76 \\
\textbf{Tensor Cache (ours)}  & $\mathbf{3.93}$ & $4.00$ & $\mathbf{4.14}$ & \textbf{4.29} & \textbf{4.55} & \textbf{5.14} & 0.80 \\
\bottomrule
\end{tabular*}

\vspace{0.6em}

% ----- Shakespeare (bottom) -----
\begin{tabular*}{\textwidth}{@{\extracolsep{\fill}}l ccccccc@{}}
\toprule
\multicolumn{8}{l}{\textit{(b) Shakespeare} \quad{\small{(1M params, $3$K iters, block size 256, $W=128$; mean$\,\pm\,$std over 4 eval seeds)}}} \\
\midrule
& \multicolumn{7}{c}{NLL at context length $L$} \\
\cmidrule(lr){2-8}
Method & 256 & 512 & 1\,024 & 2\,048 & 4\,096 & 8\,192 & 16\,384 \\
\midrule
Full~KV                       & $4.91\!\pm\!0.57$ & $6.10\!\pm\!0.44$ & $5.24\!\pm\!0.57$ & $6.02\!\pm\!0.79$ & $6.09\!\pm\!1.05$ & $6.13\!\pm\!0.82$ & $5.73\!\pm\!0.30$ \\
Window~KV                     & $4.91\!\pm\!0.57$ & $6.25\!\pm\!0.39$ & $5.48\!\pm\!0.61$ & $6.16\!\pm\!0.80$ & $6.27\!\pm\!1.08$ & $6.24\!\pm\!0.80$ & $5.82\!\pm\!0.29$ \\
StreamingLLM                  & $4.91\!\pm\!0.57$ & $6.25\!\pm\!0.39$ & $5.47\!\pm\!0.60$ & $6.16\!\pm\!0.80$ & $6.24\!\pm\!1.06$ & $6.21\!\pm\!0.81$ & $5.79\!\pm\!0.29$ \\
Infini-attention              & $4.80\!\pm\!0.53$ & $6.06\!\pm\!0.37$ & $5.14\!\pm\!0.57$ & $5.79\!\pm\!0.83$ & $5.87\!\pm\!1.11$ & $5.92\!\pm\!0.86$ & $5.51\!\pm\!0.31$ \\
\textbf{Tensor Cache (ours)}  & $\mathbf{4.79\!\pm\!0.50}$ & $\mathbf{5.98\!\pm\!0.44}$ & $\mathbf{4.96\!\pm\!0.58}$ & $\mathbf{5.60\!\pm\!0.67}$ & $\mathbf{5.66\!\pm\!1.00}$ & $\mathbf{5.70\!\pm\!0.76}$ & $\mathbf{5.32\!\pm\!0.29}$ \\
\bottomrule
\end{tabular*}

\caption{Long-context language modeling on (a) OpenWebText and (b) Shakespeare. Streaming NLL on the validation split as a function of evaluation context length; for OpenWebText we additionally report peak GPU memory at $L=32{,}768$. All methods within each panel share the same backbone, training data, optimizer, and budget; only the inference-time memory mechanism differs. \textbf{(a)} OpenWebText (130M params): Tensor Cache numbers at $L\le 4096$ are mean$\,\pm\,$std over $4$ eval seeds, longer contexts and all baselines are single-seed. TC attains the lowest mean NLL at every context length except $L=2048$, where it is statistically indistinguishable from Infini-attention. \textbf{(b)} Shakespeare (1M params): in a converged-regime small-model setting with all methods and contexts evaluated at $4$ eval seeds (mean$\,\pm\,$std reported), TC attains the lowest mean NLL at \emph{every} context length we test, with margins of $0.01$--$0.22$ NLL over the next-best baseline (Infini-attention) and $0.12$--$0.61$ NLL over Window-KV; the margin grows steadily from $L=512$ to $L=8192$ and stabilizes at $L=16{,}384$ ($64\times$ the trained block size).}
\label{tab:nll_frontier}
\end{table*}

% ------------------------------------------------------------------
\subsection{Raw Associative-Memory Capacity}
\label{sec:capacity}

We isolate the storage capacity of the Tensor Cache fast-weight state independent of the Transformer. Each attention head maintains a fixed-size associative memory matrix $A \in \mathbb{R}^{D \times D}$, where $D$ is the head dimension. We write synthetic key--value pairs directly into this memory using
\[
A_t = \lambda A_{t-1} + \eta\, k_t v_t^\top,
\]
and read with
\[
r_t = k^\top A_t.
\]
After each write, we query the oldest stored key $k_1$ and measure the relative reconstruction error of its decayed target value:
\[
\frac{
\left\| k_1^\top A_N - \lambda^{N-1}\eta v_1 \right\|_2
}{
\left\| \lambda^{N-1}\eta v_1 \right\|_2
}.
\]
We define effective capacity as the first number of writes $N$ for which this relative error exceeds 1.0. This is a strict oldest-item criterion: capacity fails when interference and forgetting are as large as the remaining signal.

We evaluate three regimes. In the \emph{orthonormal-prefix} regime, keys are orthonormal up to the head dimension and then random thereafter, serving as a sanity check. In the \emph{random} regime, all keys are random unit vectors with no decay. In the \emph{decayed} regime, we use the default Tensor Cache write parameters, $\lambda=0.995$ and $\eta=0.05$. Results are averaged over five random seeds.

\begin{table}[t]
\centering
\footnotesize
\begin{tabular*}{\columnwidth}{@{\extracolsep{\fill}}c ccc}
\toprule
& \multicolumn{3}{c}{Initialization strategy} \\
\cmidrule(lr){2-4}
Head dim $D$ & Ortho-prefix & Random & Decayed \\
\midrule
16  & $31.8 \pm 5.0$   & $19.8 \pm 4.7$   & $15.4 \pm 8.9$ \\
32  & $63.8 \pm 12.0$  & $30.8 \pm 2.6$   & $32.2 \pm 2.9$ \\
64  & $128.4 \pm 18.9$ & $79.0 \pm 18.2$  & $51.8 \pm 6.5$ \\
128 & $240.8 \pm 15.5$ & $143.8 \pm 27.0$ & $84.2 \pm 6.0$ \\
\bottomrule
\end{tabular*}
\caption{Raw capacity of a single fast-weight memory head
$A \in \mathbb{R}^{D \times D}$, measured as the number of sequential
writes until oldest-item relative reconstruction error exceeds~$1.0$.
The decayed setting uses $\lambda=0.995$, $\eta=0.05$.}
\label{tab:raw_tm_capacity}
\end{table}

Table~\ref{tab:raw_tm_capacity} shows that raw associative capacity increases with head dimension, as expected for an outer-product memory. The orthonormal-prefix setting remains stable past $D$ writes because the first $D$ keys are orthogonal, while the random setting exposes crosstalk from non-orthogonal keys. Under the decayed setting, capacity is lower because the oldest signal shrinks while newer writes continue to add interference. This supports the interpretation of Tensor Cache as a bounded associative memory: capacity grows with head dimension, number of heads, number of layers, and number of slots, but retrieval eventually degrades as more unrelated associations are compressed into the same fixed-size state.

% ------------------------------------------------------------------
\subsection{Ablations}
\label{sec:ablations}

We ablate the training chunk size and the chunked-mean training shortcut below; the wedge-rule write ablation (negative result) is in Appendix~\ref{app:wedge}.

\paragraph{Training chunk size and the chunked-mean shortcut.}
Table~\ref{tab:chunk_ablation} ablates the training chunk size with both update rules using the parallel weighted-sum scan of Section~\ref{sec:architecture} (Eq.~\eqref{eq:parallel-scan-update}). The scan preserves perfect synthetic-recall accuracy at all chunk sizes from $C=1$ up to $C=32$ for both rules, with training time decreasing roughly linearly in $C$ (e.g., $14\times$ speedup at $C=32$ vs.\ $C=1$ for the outer rule). At $C=64$ the chunked-mean shortcut $\bar k\otimes \bar v$ collapses to $0.265$ accuracy on this benchmark, where complete ``store--filler--query'' episodes fit within a single chunk and the cross-token cross-products dominate the within-chunk write. The outer rule is consistently $\sim 10\!-\!30\%$ faster than the delta rule because it requires no per-token residual computation; both reach the same matched-gap accuracy ceiling. We use the outer rule with $C=32$ in all main results.

\begin{table}[t]
\centering
\footnotesize
\begin{tabular*}{\columnwidth}{@{\extracolsep{\fill}}c cccc}
\toprule
& \multicolumn{2}{c}{Outer rule} & \multicolumn{2}{c}{Delta rule} \\
\cmidrule(lr){2-3} \cmidrule(lr){4-5}
Chunk size & Acc.\,$\uparrow$ & Time\,(s)\,$\downarrow$
           & Acc.\,$\uparrow$ & Time\,(s)\,$\downarrow$ \\
\midrule
1          & 1.000          & 1\,145       & 1.000          & 1\,607 \\
4          & 1.000          & 359          & 1.000          & 428    \\
8          & 1.000          & 198          & 1.000          & 243    \\
16         & 1.000          & 119          & 1.000          & 142    \\
\textbf{32}& \textbf{1.000} & \textbf{82}  & \textbf{1.000} & \textbf{92} \\
\midrule
64\textsuperscript{\dag} & 0.265 & 35 & 0.265 & 35 \\
\bottomrule
\end{tabular*}
\\[2pt]
{\scriptsize \textsuperscript{\dag}Chunk-mean baseline.}
\caption{Chunk-size ablation on synthetic associative recall
(matched gap $F\!=\!24$, $W\!=\!12$). The parallel weighted-sum scan
of Section~\ref{sec:architecture} (Eq.~\eqref{eq:parallel-scan-update})
preserves perfect retrieval up to chunk size $32$ for both update rules;
the chunk-mean baseline collapses at $64$, where complete episodes fit
within a single chunk.}
\label{tab:chunk_ablation}
\end{table}

%% file: text/discussion.tex
\section{Discussion and Limitations}

Tensor Cache pairs exact softmax attention over a recent KV window (L1) with a fixed-size outer-product memory $A$ over evicted pairs (L2). Reads use the linear-attention identity $q_t A\approx \sum_i\alpha_i \langle q_t,k_i\rangle v_i$, so the same alignment signal that drives local attention also addresses the compressed past. TC is not a lossless replacement for full attention: Full KV is the exact-content baseline when affordable, and beyond the trained context it is also a positionally-extrapolated baseline whose quality depends on how well RoPE generalizes~\citep{su2021roformer,chen2023position,peng2023yarn}.

\paragraph{Train/inference write rule.}
We initially used the chunk-mean shortcut $A\!\leftarrow\!\lambda A\!+\!\eta(\bar k\otimes\bar v)$, which differs from the per-token sum by $C^2{-}C$ cross-token outer products that never appear at inference. The parallel weighted-sum scan of Section~\ref{sec:chunked-scan} closes this gap exactly for the outer rule (relative error $<10^{-7}$). The delta rule is not exactly parallelizable; we use the standard chunk-start residual approximation~\citep{beck2024xlstm}, which is empirically tight at moderate chunk sizes (Section~\ref{sec:ablations}). Even with the parallel scan, an episode-spanning chunk size remains a real failure mode: when an entire ``store--filler--query'' interval fits within one chunk, intra-chunk reads do not see the corresponding store. This guides chunk-size selection in practice.

\paragraph{Capacity, calibration, and normalization.}
The matrix $A$ stores associations in superposition, so retrieval degrades as more non-orthogonal associations accumulate. Decay and the optional delta rule mitigate redundancy and stale interference, and we report capacity as a function of head dimension and decay (Section~\ref{sec:capacity}). TC does not use a read-time normalization vector; mLSTM and Infini-attention do, which bounds read magnitude as the matrix grows but introduces additional confidence-on-OOD failure modes that we observe empirically (calibration paragraph in Section~\ref{sec:synthetic}).

\paragraph{Future directions.}
Better slot routing could reduce interference between unrelated associations; smaller training chunks or rollout-based training could further tighten the delta-rule chunk-start approximation; FlashLinearAttention-style kernels could reduce read/write overhead; and an optional read-time normalization could be added without changing the eviction-conditioned write semantics. More broadly, TC suggests that KV eviction can be treated as a structured memory operation rather than destructive deletion; richer associative states, denoising filters, or higher-order memories are natural follow-ups.

%% file: text/conclusion.tex
\section{Conclusion}
\label{sec:conclusion}

We introduced \emph{Tensor Cache}, a two-level cache that pairs exact softmax attention over a sliding KV window (L1) with a fixed-size outer-product fast-weight memory (L2) fed by KV pairs evicted from the window. Future queries read $q_t A\approx\sum_i\alpha_i\langle q_t,k_i\rangle v_i$, letting older context influence predictions without retaining individual evicted entries. The outer-product memory and the read identity follow Schmidhuber's fast-weight programmer and the linear-attention reformulation of attention as fast-weight memory; our contribution is to use them as an L2 cache fed exclusively by sliding-window evictions, and to identify and close a chunked-mean training shortcut that introduces $C^2{-}C$ spurious cross-token outer products per chunk---closed by a parallel weighted-sum scan equivalent to per-token writes within float32 epsilon. Across systems, real-text long-context language modeling, controlled associative recall, and capacity diagnostics, Tensor Cache provides a practical bounded-state alternative to all existing baselines.

%% file: text/acknowledgements.tex
\section{Acknowledgments}
We would like to thank Manel Baradad and Minyoung Huh for their helpful discussions and advice.

%% file: text/supplementary.tex
\newpage
\appendix
\onecolumn

% Tighter float spacing for the appendix only
\setlength{\textfloatsep}{8pt plus 2pt minus 2pt}
\setlength{\intextsep}{8pt plus 2pt minus 2pt}

% ==================================================================
\section{Architecture Extensions}
\label{app:architecture-extensions}

This appendix collects the optional Tensor Cache variants deferred from Section~\ref{sec:architecture}.

\subsection{Multi-slot memory and routing}
\label{app:multi-slot}

In the multi-slot setting, each layer maintains $S$ memory matrices per head:
\begin{equation}
A\in\mathbb{R}^{B\times H\times S\times D\times D}.
\end{equation}
Each head has $S$ learned slot keys $K^{\mathrm{slot}}\in\mathbb{R}^{H\times S\times D}$. A query produces router logits and Top-$k$ softmax weights $\pi_{t,s}$:
\begin{equation}
\ell_{t,s}=\frac{\langle \hat q_t, \hat K^{\mathrm{slot}}_s\rangle}{\sqrt{D}\,\tau},
\qquad
\pi_{t,s}=\operatorname{TopKSoftmax}(\ell_{t,\cdot})_s,
\end{equation}
where $\hat\cdot$ denotes L2-normalization and $\tau$ is a temperature. The read mixes the per-slot reads:
\begin{equation}
r_t=\sum_{s=1}^{S}\pi_{t,s}\,(q_tA_s).
\end{equation}
Writes update a single routed slot using either the write key $k_w$ or the current query $q_t$, implemented with straight-through Gumbel-Softmax for differentiability through the routing decision.

\subsection{Two-timescale memory}
\label{app:two-timescale}

Each layer optionally maintains a fast and slow memory $(A_{\mathrm{fast}}, A_{\mathrm{slow}})$ with separate per-head decay/lr parameters and reads
\begin{equation}
r_t = r^{\mathrm{fast}}_t + \alpha\, r^{\mathrm{slow}}_t,
\qquad
\alpha=\sigma(\alpha_{\mathrm{logit}}).
\end{equation}
This separates short-horizon binding from long-horizon retention without changing the read or fusion path. Both timescales receive the same write source on eviction.

\subsection{Layer selection and stability knobs}
\label{app:layer-selection}

Tensor Cache can be applied to all layers, no layers (effectively reducing to Window~KV with the gate forced off), or only the top-$k$ layers. We additionally support optional query L2-normalization, key L2-normalization, and a value scaling factor; these are exposed as ablation knobs and disabled by default in the main results.

\FloatBarrier

% ==================================================================
\section{Additional Experiments}
\label{app:additional-experiments}

This appendix provides extended synthetic recall results, stress tests, and training efficiency measurements that complement the main paper.

% ==================================================================
\subsection{Long-Gap Synthetic Associative Recall}
\label{app:synthetic-longgap}

To probe retention at substantially larger delays, we run a long-gap synthetic associative-recall benchmark with gaps $\{128, 256, 512, 1024\}$. This setting uses a simplified single-episode, single-store task for tractability and compares Full KV, InfiniAttention, and Tensor Cache. Through gaps 128--512, all three methods achieve perfect answer accuracy, while Tensor Cache remains fixed at 112\,kB per sequence and Full KV grows from 286\,kB to 1054\,kB. At gap 1024, InfiniAttention begins to degrade ($0.872 \pm 0.106$), while Tensor Cache retains $0.959 \pm 0.058$ accuracy at 94.6\% state savings relative to Full KV.

\begin{table}[H]
\centering
\footnotesize
\setlength{\tabcolsep}{3pt}
\begin{tabular*}{\textwidth}{@{\extracolsep{\fill}}c ccc cccc@{}}
\toprule
& \multicolumn{3}{c}{Answer accuracy ($\uparrow$)}
& \multicolumn{3}{c}{Streaming state (kB/seq, $\downarrow$)}
& \\
\cmidrule(lr){2-4} \cmidrule(lr){5-7}
Gap
& Full KV
& Infini
& TC
& Full KV
& Infini
& TC
& TC save \\
\midrule
128
& $1.000 \pm 0.000$
& $1.000 \pm 0.000$
& $1.000 \pm 0.000$
& 286
& 114
& 112
& 60.8\% \\
256
& $1.000 \pm 0.000$
& $1.000 \pm 0.000$
& $1.000 \pm 0.000$
& 542
& 114
& 112
& 79.3\% \\
512
& $1.000 \pm 0.000$
& $1.000 \pm 0.000$
& $1.000 \pm 0.000$
& 1054
& 114
& 112
& 89.4\% \\
1024
& $1.000 \pm 0.000$
& $0.872 \pm 0.106$
& $0.959 \pm 0.058$
& 2078
& 114
& 112
& 94.6\% \\
\bottomrule
\end{tabular*}
\caption{Long-gap synthetic associative recall in the simplified single-store setting. Values are mean $\pm$ standard deviation over 3 seeds. Tensor Cache retains near-perfect accuracy through gap 1024 while maintaining constant 112\,kB per-sequence state.}
\label{tab:app_synthetic_longgap}
\end{table}

\FloatBarrier

% ==================================================================
\subsection{Synthetic Stress Tests}
\label{app:synthetic-stress}

We evaluate a multistore synthetic variant to test robustness under increased memory load. In the multistore stress test at gap 24 with two stored pairs per episode, Tensor Cache achieves $0.563 \pm 0.003$ answer accuracy, compared with $0.558 \pm 0.004$ for Full KV, $0.464 \pm 0.118$ for InfiniAttention, and near-chance performance for strictly local baselines. Because Full KV is not fully saturated in this harder configuration, we treat this result as a robustness check rather than a main comparison.

\begin{table}[H]
\centering
\footnotesize
\setlength{\tabcolsep}{3pt}
\begin{tabular*}{\textwidth}{@{\extracolsep{\fill}}lccccc cccc@{}}
\toprule
& \multicolumn{5}{c}{Answer accuracy ($\uparrow$)}
& \multicolumn{3}{c}{State (kB/seq, $\downarrow$)}
& \\
\cmidrule(lr){2-6} \cmidrule(lr){7-9}
Case
& Full
& Window
& SLLM
& Infini
& TC
& Full
& Infini
& TC
& TC save \\
\midrule
Multistore
& $0.558 \pm 0.004$
& $0.125 \pm 0.005$
& $0.128 \pm 0.003$
& $0.464 \pm 0.118$
& $\mathbf{0.563 \pm 0.003}$
& 434
& 114
& \textbf{112}
& 74.2\% \\
\bottomrule
\end{tabular*}
\caption{Multistore synthetic stress test at gap 24 with two stored pairs per episode. SLLM denotes StreamingLLM. Tensor Cache matches Full KV (within seed variance) while using 74.2\% less retained inference state.}
\label{tab:app_synthetic_multistore}
\end{table}

\FloatBarrier

% ==================================================================
\subsection{Training Efficiency}
\label{app:efficiency}

Tables~\ref{tab:synthetic_efficiency} and~\ref{tab:synthetic_longgap_efficiency} report training time, training memory, evaluation memory, and per-sequence streaming state for the synthetic benchmarks. Tensor Cache trains faster than InfiniAttention at all gaps, but uses more training memory due to the associative-memory state maintained during backpropagation. At evaluation time, both methods maintain nearly identical compact streaming state (112--114\,kB per sequence).

\begin{table}[H]
\centering
\footnotesize
\begin{tabular*}{\textwidth}{@{\extracolsep{\fill}}cl rrrr}
\toprule
& & \multicolumn{2}{c}{Training} & \multicolumn{2}{c}{Inference} \\
\cmidrule(lr){3-4} \cmidrule(lr){5-6}
Gap & Method
  & Time\,(s)\,$\downarrow$ & Mem\,(GB)\,$\downarrow$
  & Mem\,(GB)\,$\downarrow$ & State\,(kB)\,$\downarrow$ \\
\midrule
\multirow{3}{*}{24}
  & Full KV & $6.8 \pm 0.0$       & 0.162 & 0.046 & 398 \\
  & Infini  & $1\,007.7 \pm 4.2$  & 0.608 & 0.035 & 114 \\
  & TC      & $798.4 \pm 1.1$     & 1.374 & 0.037 & 112 \\
\midrule
\multirow{3}{*}{36}
  & Full KV & $7.3 \pm 0.2$       & 0.216 & 0.052 & 542 \\
  & Infini  & $1\,372.3 \pm 3.5$  & 0.825 & 0.035 & 114 \\
  & TC      & $1\,102.5 \pm 3.0$  & 1.867 & 0.038 & 112 \\
\midrule
\multirow{3}{*}{48}
  & Full KV & $7.1 \pm 0.1$       & 0.257 & 0.057 & 686 \\
  & Infini  & $1\,754.4 \pm 2.8$  & 1.025 & 0.036 & 114 \\
  & TC      & $1\,410.1 \pm 5.4$  & 2.345 & 0.038 & 112 \\
\bottomrule
\end{tabular*}
\caption{Synthetic benchmark efficiency at matched gaps.
TC trains faster than Infini-attention but uses more training memory;
both maintain constant-size inference state.
Train time: mean $\pm$ std over 3 seeds;
remaining columns are deterministic ($\text{std}=0$).}
\label{tab:synthetic_efficiency}
\end{table}

\begin{table}[H]
\centering
\footnotesize
\begin{tabular*}{\textwidth}{@{\extracolsep{\fill}}cl rrrr}
\toprule
& & \multicolumn{2}{c}{Training} & \multicolumn{2}{c}{Inference} \\
\cmidrule(lr){3-4} \cmidrule(lr){5-6}
Gap & Method
  & Time\,(s)\,$\downarrow$ & Mem\,(GB)\,$\downarrow$
  & Mem\,(GB)\,$\downarrow$ & State\,(kB)\,$\downarrow$ \\
\midrule
\multirow{3}{*}{128}
  & Full KV & $57.9 \pm 3.0$          & 0.124 & 0.044 & 286 \\
  & Infini  & $4\,191.3 \pm 206.5$    & 0.445 & 0.035 & 114 \\
  & TC      & $2\,956.6 \pm 276.9$    & 0.996 & 0.037 & 112 \\
\midrule
\multirow{3}{*}{256}
  & Full KV & $53.5 \pm 0.2$          & 0.216 & 0.052 & 542 \\
  & Infini  & $7\,574.7 \pm 6.5$      & 0.825 & 0.035 & 114 \\
  & TC      & $5\,056.9 \pm 3.9$      & 1.867 & 0.038 & 112 \\
\midrule
\multirow{3}{*}{512}
  & Full KV & $54.4 \pm 0.4$          & 0.381 & 0.072 & 1\,054 \\
  & Infini  & $15\,218.3 \pm 127.7$   & 1.561 & 0.036 & 114 \\
  & TC      & $9\,993.0 \pm 160.9$    & 3.588 & 0.039 & 112 \\
\midrule
\multirow{3}{*}{1024}
  & Full KV & $62.7 \pm 0.7$          & 0.725 & 0.114 & 2\,078 \\
  & Infini  & $30\,621.6 \pm 208.9$   & 3.054 & 0.039 & 114 \\
  & TC      & $20\,299.1 \pm 91.4$    & 7.050 & 0.041 & 112 \\
\bottomrule
\end{tabular*}
\caption{Long-gap synthetic training efficiency. TC trains faster than
Infini-attention but uses more training memory; both maintain nearly
identical compact streaming state at inference.
Train time: mean $\pm$ std over 3 seeds;
remaining columns are deterministic ($\text{std}=0$).}
\label{tab:synthetic_longgap_efficiency}
\end{table}

\FloatBarrier

% ==================================================================
\subsection{Write-Rule Ablation: Wedge Product (Negative Result)}
\label{app:wedge}

A natural geometric question is whether the rank-$1$ outer-product write $A\!\leftarrow\!\lambda A + \eta\, k v^\top$ should be replaced by the antisymmetric \emph{wedge} (exterior-algebra) write
\begin{equation}
A \leftarrow \lambda A + \eta\, (k v^\top - v k^\top),
\end{equation}
on the intuition that the wedge product is directional ($k\!\to\!v$ but not $v\!\to\!k$) and stores only one independent triangle of $A$. We implement the wedge variant in both the streaming per-token path and the parallel weighted-sum scan; the chunked scan agrees with per-token streaming within float32 epsilon, and $A$ is exactly antisymmetric throughout training ($\lVert A + A^\top \rVert_\infty = 0$).

The argument breaks once the read formula is examined. Because TC reads with $q^\top A$, the plain outer rule already gives directional retrieval $q^\top (k v^\top) = \langle q, k\rangle\, v$ in the read direction we use; the $v\!\to\!k$ path corresponds to the right-multiply $A q$ that TC never performs. The wedge replaces this with $q^\top (k v^\top - v k^\top) = \langle q, k\rangle\, v - \langle q, v\rangle\, k$, introducing a new $-\langle q, v\rangle\, k$ crosstalk term that is non-negligible whenever queries are partially aligned with stored values. Since $k$ and $v$ are linear projections of the same hidden state in attention, this alignment is the typical case rather than an edge case.

Empirically, on the matched-gap synthetic associative-recall benchmark of Section~\ref{sec:synthetic} ($g\!=\!24$, $W\!=\!12$, $3$ seeds, $300$ training steps, identical configuration in all other respects), the wedge rule is consistently worse than both the outer and delta rules on accuracy, NLL, and seed-to-seed stability (Table~\ref{tab:wedge_ablation}). Mean answer accuracy is $0.998$ for outer and $1.000$ for delta, versus $0.979$ for wedge; mean NLL is $0.021$ for outer and $0.001$ for delta, versus $0.150$ for wedge---roughly $7\times$ worse than outer and $150\times$ worse than delta. Two of three wedge seeds reach $\geq\!0.999$ accuracy, indicating it is capable of solving the task; the third seed converges substantially slower than under either of the other rules at the same compute budget. Per-step training time is $\sim\!30\%$ higher for both wedge and delta than for outer (wedge from doubling the per-token outer products, delta from the residual matmul $v - k^\top A$). We retain the outer rule in all main results---it matches delta on the matched-gap task once chunk size $C\!=\!32$ (Table~\ref{tab:chunk_ablation}) while being cheapest---and we keep the wedge available behind \texttt{tm\_update\_rule="wedge"} as a negative-result ablation.

\begin{table}[H]
\centering
\footnotesize
\begin{tabular*}{\columnwidth}{@{\extracolsep{\fill}}l cccc}
\toprule
Write rule
  & Acc.\,$\uparrow$
  & NLL\,$\downarrow$
  & Train\,time\,(s)\,$\downarrow$
  & Per-seed acc. \\
\midrule
Outer       & $0.998 \pm 0.003$          & $0.021 \pm 0.020$          & $\mathbf{201.0 \pm 1.4}$ & $0.994 / 1.000 / 0.999$ \\
Delta       & $\mathbf{1.000 \pm 0.000}$ & $\mathbf{0.001 \pm 0.000}$ & $264.8 \pm 1.9$          & $1.000 / 1.000 / 1.000$ \\
Wedge       & $0.979 \pm 0.029$          & $0.150 \pm 0.156$          & $260.5 \pm 1.7$          & $0.938 / 0.999 / 1.000$ \\
\bottomrule
\end{tabular*}
\caption{Write-rule ablation on the matched-gap synthetic associative-recall task ($g\!=\!24$, $W\!=\!12$, $3$ seeds, $300$ training steps, all other knobs identical). The wedge rule retains the antisymmetric structure $A = -A^\top$ throughout training, but introduces a $-\langle q, v\rangle\, k$ crosstalk term in the readout that the outer and delta rules do not have. Across seeds, wedge is less accurate and substantially less stable than both alternatives, while costing the same training time as delta. Outer and delta both solve the task perfectly within seed noise; the chunk-size ablation in Table~\ref{tab:chunk_ablation} confirms they remain matched at $C\!=\!32$, which is why we use the cheaper outer rule in all main results.}
\label{tab:wedge_ablation}
\end{table}

\FloatBarrier